\documentclass[10pt,twocolumn,letterpaper]{article}

\usepackage{titling}
\usepackage{cvpr}
\usepackage{times}
\usepackage{epsfig}
\usepackage{graphicx}
\usepackage{amsmath}
\usepackage{amssymb}
\usepackage{booktabs}
\usepackage{subcaption}
\usepackage{multirow}
\usepackage{comment}
\usepackage{tabu,booktabs}

\graphicspath{{figures/}}

\DeclareMathOperator*{\argmax}{arg\,max}

\usepackage[breaklinks=true,bookmarks=false]{hyperref}

\cvprfinalcopy 

\def\cvprPaperID{1386} 

\begin{document}

\title{Learning in the Frequency Domain}

\author{Kai Xu$^{1,2}$\thanks{Work partially done during an internship at Alibaba.}
\quad\quad\quad\quad\quad
Minghai Qin$^{1}$
\quad\quad\quad\quad
Fei Sun$^{1}$ \\
\ \ \
Yuhao Wang$^{1}$
\quad\quad\quad
Yen-kuang Chen$^{1}$
\quad\quad\quad
Fengbo Ren$^{2}$
\vspace{.5em} \\
$^1$DAMO Academy, Alibaba Group \qquad $^2$Arizona State University
}

\maketitle

\begin{abstract}
	Deep neural networks have achieved remarkable success in computer vision tasks.
	Existing neural networks mainly operate in the spatial domain with fixed input sizes. 
	For practical applications, images are usually large and have to be downsampled to the predetermined input size of neural networks. Even though the downsampling operations reduce computation and the required communication bandwidth, it removes both redundant and salient information obliviously, which results in accuracy degradation. Inspired by digital signal processing theories, we analyze the spectral bias from the frequency perspective and propose a learning-based frequency selection method to identify the trivial frequency components which can be removed without accuracy loss.
	The proposed method of learning in the frequency domain leverages identical structures of the well-known neural networks, such as ResNet-50, MobileNetV2, and Mask R-CNN, while accepting the frequency-domain information as the input.
	Experiment results show that learning in the frequency domain with static channel selection can achieve higher accuracy than the conventional spatial downsampling approach and meanwhile further reduce the input data size. Specifically for ImageNet classification with the same input size, the proposed method achieves $1.60\%$ and $0.63\%$ top-1 accuracy improvements on ResNet-50 and MobileNetV2, respectively. Even with half input size, the proposed method still improves the top-1 accuracy on ResNet-50 by $1.42\%$. In addition, we observe a $0.8\%$ average precision improvement on Mask R-CNN for instance segmentation on the COCO dataset.
\end{abstract}

\begin{figure*}[ht]
	\centering
	\begin{subfigure}{0.75\linewidth}
		\centering
		\includegraphics[width=\textwidth]{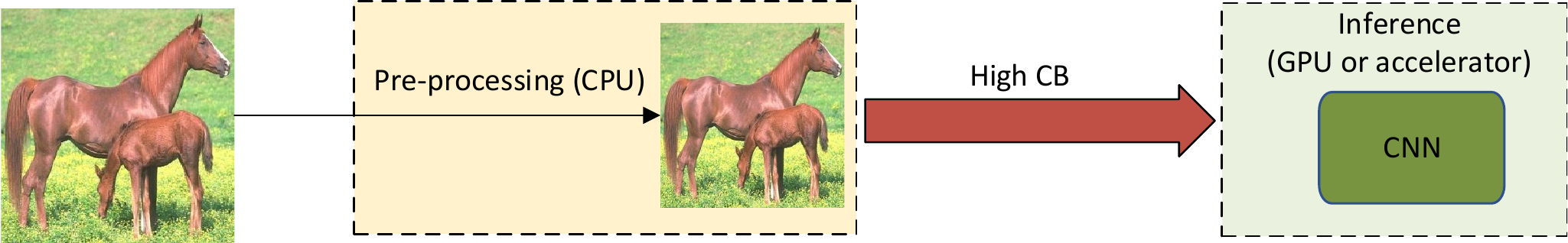}
		\caption{}
	\end{subfigure}
	\\
	\begin{subfigure}{0.75\linewidth}
		\centering
		\includegraphics[width=\textwidth]{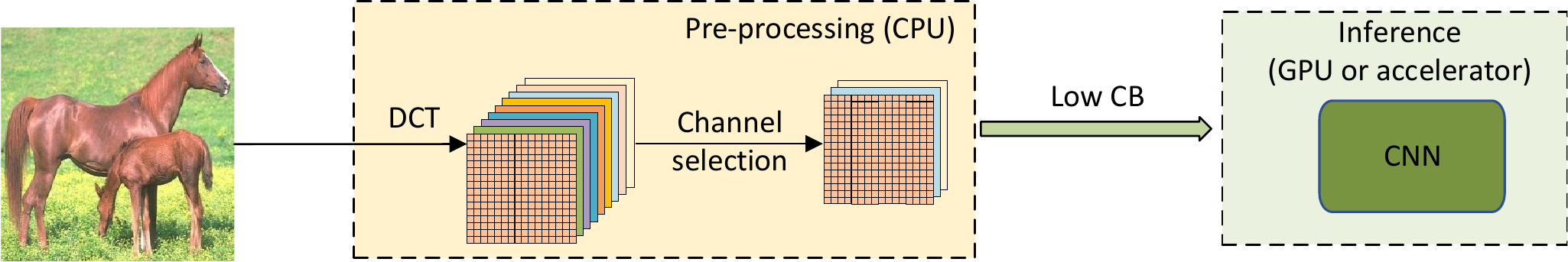}
		\caption{}
	\end{subfigure}
	\caption{(a) The workflow of the conventional CNN-based methods using RGB images as input. (b) The workflow of the proposed method using DCT coefficients as input. CB represents the required communication bandwidth between CPU and GPU/accelerator.}
	\label{fig:overall}
	\vspace{-1ex}
\end{figure*}

\section{Introduction}
Convolutional neural networks (CNNs) have revolutionized the computer vision community because of their exceptional performance on various tasks such as image classification~\cite{alex2012alexnet, karpathy2014large}, object detection~\cite{ren2015faster, redmon2016yolo}, and semantic segmentation~\cite{long2015fully, chen2018encoder}. 
Constrained by the computing resources and memory limitations, most CNN models only accept RGB images at low resolutions (\textit{e.g.}, $224\times224$). However, images produced by modern cameras are usually much larger. For example, the high definition (HD) resolution images ($1920$$\times$$1080$) are considered relatively small by modern standards. Even the average image resolution in the ImageNet dataset~\cite{olga2015imagenet} is 482$\times$415, which is roughly four times the size accepted by most CNN models. Therefore, a large portion of real-world images are aggressively downsized to $224$$\times$$224$ to meet the input requirement of classification networks. However, image downsizing inevitably incurs information loss and accuracy degradation \cite{pei2019effects}. 
Prior works \cite{kim2018task, saeedan2018dpp} aim to reduce information loss by learning task-aware downsizing networks. However, those networks are task-specific and require additional computation, which are not favorable in practical applications. In this paper, we propose to reshape the high-resolution images in the frequency domain, \textit{i.e.}, discrete cosine transform (DCT) domain \footnote{ We interchangeably use the terms frequency domain and DCT domain in the context of this paper.}, rather than resizing them in the spatial domain, and then feed the reshaped DCT coefficients to CNN models for inference. Our method requires little modification to the existing CNN models that take RGB images as input. Thus, it is a universal replacement for the routine data pre-processing pipelines. We demonstrate that our method achieves higher accuracy in image classification, object detection, and instance segmentation tasks than the conventional RGB-based methods with an equal or smaller input data size. The proposed method leads to a direct reduction in the required inter-chip communication bandwidth that is often a bottleneck in modern deep learning inference systems, \textit{i.e.}, the computational throughput of rapidly evolving AI accelerators/GPUs is becoming increasingly higher than the data loading throughput of CPUs, as shown in Figure~\ref{fig:overall}.

Inspired by the observation that human visual system (HVS) has unequal sensitivity to different frequency components~\cite{kim2017deep}, we analyze the image classification, detection and segmentation task in the frequency domain and find that CNN models are more sensitive to low-frequency channels than the high-frequency channels, which coincides with HVS. 
This observation is validated by a learning-based channel selection method that consists of multiple ``on-off switches".
The DCT coefficients with the same frequency are packed as one channel, and each switch is stacked on a specific frequency channel to either allow the entire channel to flow into the network or not. 


Using the decoded high-fidelity images for model training and inference has posed significant challenges, from both data transfer and computation perspectives  \cite{wei2019overcoming, you2018imagenet}. 
Due to the spectral bias of the CNN models, one can only keep the important frequency channels during inference without losing accuracy. In this paper, we also develop a static channel selection approach to preserve the salient channels rather than using the entire frequency spectrum for inference. Experiment results show that the CNN models still retain the same accuracy when the input data size is reduced by 87.5\%.


The contributions of this paper are as follows:
\vspace{-1ex}
\begin{itemize}
	\itemsep 0em 
	\item We propose a method of learning in the frequency domain (using DCT coefficients as input), which requires little modification to the existing CNN models that take RGB input. We validate our method on ResNet-50 and MobileNetV2 for the image classification task and Mask R-CNN for the instance segmentation task.
	\item We show that learning in the frequency domain better preserves image information in the pre-processing stage than the conventional spatial downsampling approach (spatially resizing the images to $224$$\times$$224$, the default input size of most CNN models) and consequently achieves improved accuracy, \textit{i.e.}, $+1.60\%$ on ResNet-50 and $+0.63\%$ on MobileNetV2 for the ImageNet classification task, $+0.8\%$ on Mask R-CNN for both object detection and instance segmentation tasks.
	\item We analyze the spectral bias from the frequency perspective and show that the CNN models are more sensitive to low-frequency channels than high-frequency channels, similar to the human visual system (HVS). 
	\item We propose a learning-based dynamic channel selection method to identify the trivial frequency components for static removal during inference. Experiment results on ResNet-50 show that one can prune up to $87.5\%$ of the frequency channels using the proposed channel selection method with no or little accuracy degradation in the ImageNet classification task.
	\item To the best of our knowledge, this is the first work that explores learning in the frequency domain for object detection and instance segmentation. Experiment results on Mask R-CNN show that learning in the frequency domain can achieve a $0.8\%$ average precision improvement for the instance segmentation task on the COCO dataset. 
\end{itemize}

\begin{figure*} [htp]
	\centering
	\includegraphics[width=0.81\textwidth]{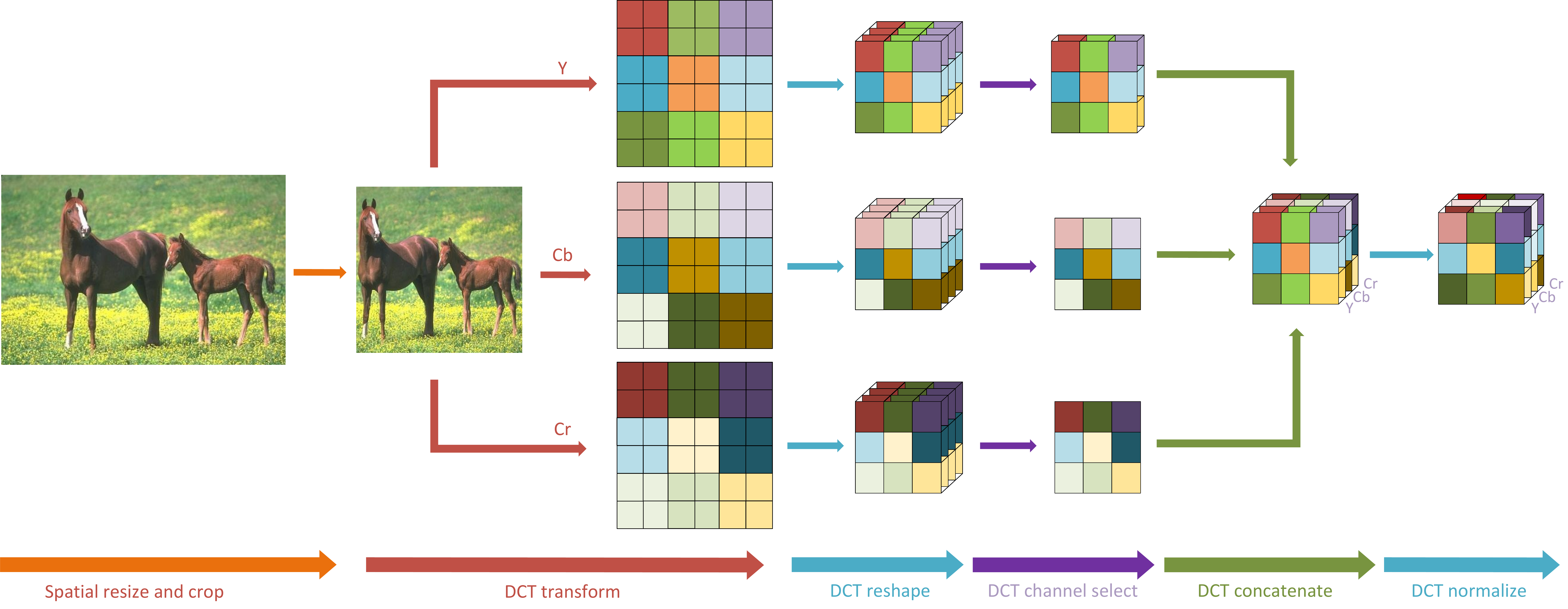}
	\caption{The data pre-processing pipeline for learning in the frequency domain.}
	\label{fig:preprocessing}
	\vspace{-1ex}
\end{figure*}

\section{Related Work}

\noindent
\textbf{Learning in the frequency domain:}  Compressed representations in the frequency domain contain rich patterns for image understanding tasks. \cite{torfason2018towards, xu2018lapran, wu2018coviar} train dedicated autoencoder-based networks on compression and inference tasks jointly. 
~\cite{gueguen2018jpeg} extracts features from the frequency domain to classify images. ~\cite{ehrlich2019jpeg} proposes a model conversion algorithm to convert the spatial-domain CNN models to the frequency domain. Our method differs from the prior works in two aspects. First, we avoid the complex model transition procedure from the spatial to the frequency domain. Thus, our method has a broader application scope. Second, we provide an analysis method to interpret the spectral bias of neural networks in the frequency domain.

\noindent
\textbf{Dynamic Neural Networks:} Prior works~\cite{veit2018aig, wang2018skipnet, guo2019dynamic, wu2018block, chen2019you} propose to selectively skip the convolutional blocks on the fly based on the activations of the previous blocks. These works adjust the model complexity in response to the input of each convolutional block. Only the intermediate features that are most relevant to the inputs are computed in the inference stage to reduce computation cost. In contrast, our method exclusively operates on the raw inputs and distills the salient frequency components to lower the communication bandwidth requirement for input data.

\noindent
\textbf{Efficient Network Training:} There are substantial recent interests in training efficient networks~\cite{frankle2018the, molchanov2019importance, wang2019haq, han2015deep}, which focus on network compression via kernel pruning, learned quantization, and entropy encoding. Another line of works aim to compress the CNN models in the frequency domain. 
~\cite{chen2016ccn} reduces the storage space by converting filter weights to the frequency domain and using a hash function to group the frequency parameters into hash buckets.~\cite{wang2019packing} also transforms the kernels to the frequency domain and discards the low-energy frequency coefficients for high compression.
~\cite{dziedzic2019band} constrains the frequency spectra of CNN kernels to reduce memory consumption. These network compression works in the frequency domain all rely on the FFT-based convolution, which is generally more effective on larger kernels. Nevertheless, the state-of-the-art CNN models use small kernels, {\it e.g.}, $3\times3$ or $1\times1$. Extensive efforts need to be taken to optimize the computation efficiency of these FFT-based CNN models~\cite{lavin2016fast}. In contrast, our method makes little modification to the existing CNN models. Thus, our method requires no extra effort to improve its computation efficiency on the CNN models with small kernels. Another fundamental difference is that our method aims at reducing the input data size rather than model complexity.

\section{Methodology}
In this paper, we propose a generic method on learning in the frequency domain, including a data pre-processing pipeline as well as an input data size pruning method. 

Figure~\ref{fig:overall} shows the comparison of our method and the conventional approach. In the conventional approach, high-resolution RGB images are usually pre-processed on a CPU and transmitted to a GPU/AI accelerator for real-time inference. Because uncompressed images in the RGB format are usually large, the requirement of the communication bandwidth between a CPU and a GPU/AI accelerator is usually high. Such communication bandwidth can be the bottleneck of the system performance, as shown in Figure~\ref{fig:overall}(a). To reduce both the computation cost and the communication bandwidth requirement, high-resolution RGB images are downsampled to smaller images, which often results in information loss and thus lower inference accuracy.

In our method, high-resolution RGB images are still pre-processed on a CPU. However, they are first transformed to the YCbCr color space and then to the frequency domain. This coincides with the most widely-used image compression standards, such as JPEG. All components of the same frequency  are grouped into one channel. In this way, multiple frequency channels are generated. As shown in Section~\ref{sec:selection}, certain frequency channels have bigger impact on the inference accuracy than the others. Thus, we propose to only preserve and transmit the most important frequency channels to a GPU/AI accelerator for inference. Compared to the conventional approach, the proposed method requires less communication bandwidth and achieves higher accuracy at the same time.


We demonstrate that the input features in the frequency domain can be applied to all existing CNN models developed in the spatial domain with minimal modification. Specifically, one just need to remove the input CNN layer and reserve the remaining residual blocks. The first residual layer is used as the input layer, and the number of input channels is modified to fit the dimension of the DCT coefficient inputs. As such, a modified model can maintain similar parameter count and computational complexity to the original model.

Based on our frequency-domain model, we propose a learning-based channel selection method to explore the spectral bias of a given CNN model, \textit{i.e.}, which frequency components are more informative to the subsequent inference task. The findings motivate us to prune the trivial frequency components for inference, which significantly reduces the input data size, consequently reducing both the computational complexity of domain transformation and the required communication bandwidth, while maintaining inference accuracy.

\begin{figure} [t]
	\centering
	\includegraphics[width=0.24\textwidth]{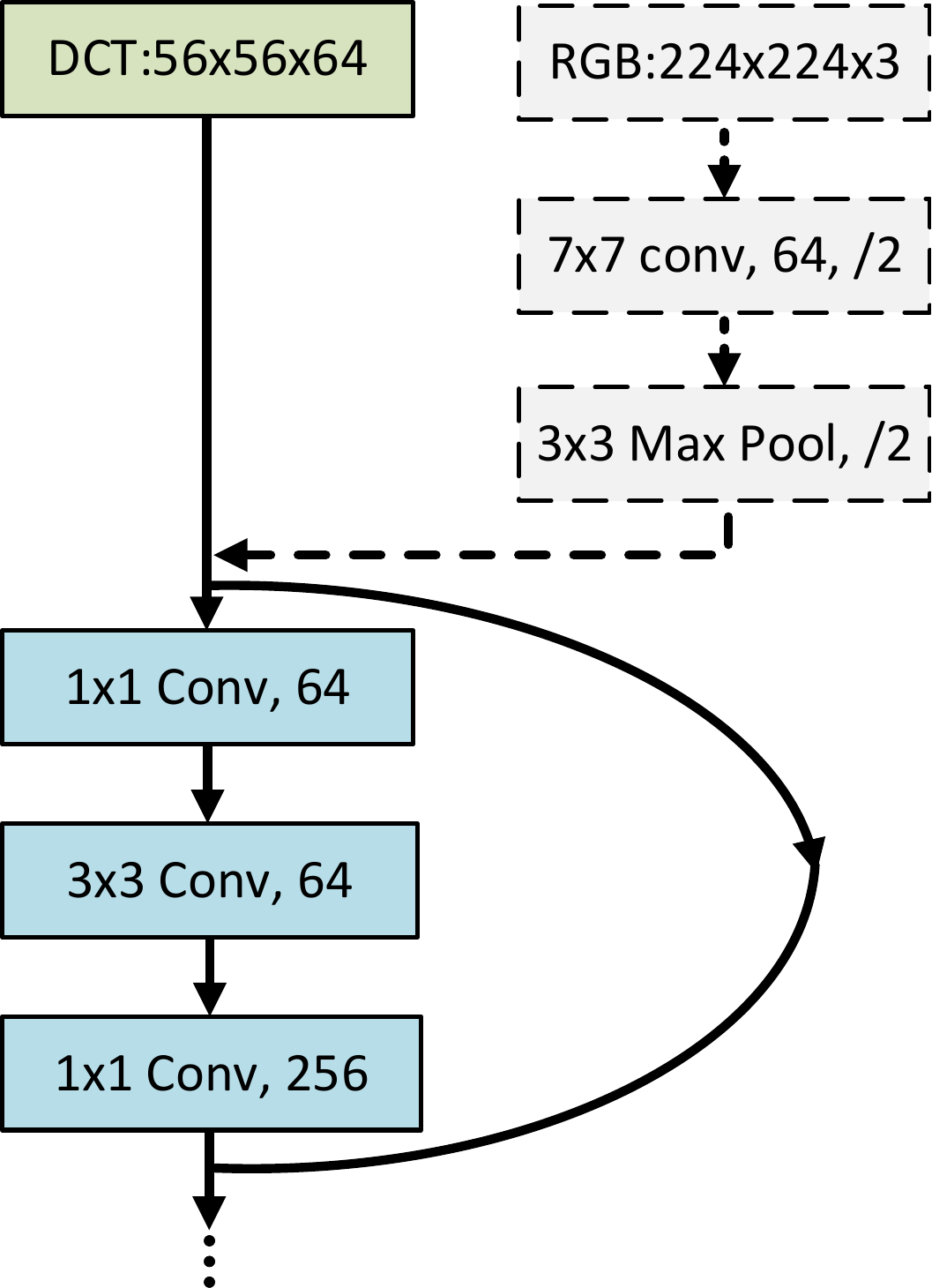}
	\caption{Connecting the pre-processed input features in the frequency domain to ResNet-50. The three input layers (the dashed gray blocks) in a vanilla ResNet-50 are removed to admit the 56$\times$56$\times$64 DCT inputs. We take $64$ channels as an example. This value can vary based on the channel selection. In learning-based channel selection, all $192$ channels are analyzed for their importance to accuracy, based on which only a subset ($\ll192$ channels) is used in the static selection approach.
	}
	\label{fig:connection}
	\vspace{-2ex}
\end{figure}

\subsection{Data Pre-processing in the Frequency Domain}\label{sec:preprocess}
The data pre-processing flow is shown in Figure~\ref{fig:preprocessing}. We follow the pre-processing and augmentation flow in the spatial domain, consisting of image resizing, cropping, and flipping (spatial resize and crop in Figure~\ref{fig:preprocessing}). Then images are transformed to the YCbCr color space and converted to the frequency domain (DCT transform in Figure~\ref{fig:preprocessing}). The two-dimensional DCT coefficients at the same frequency are grouped into one channel to form three-dimensional DCT cubes (DCT reshape in Figure~\ref{fig:preprocessing}). As will be discussed in Section \ref{sec:selection}, a subset of impactful frequency channels are selected (DCT channel select in Figure~\ref{fig:preprocessing}). The selected channels in the YCbCr color space are concatenated together to form one tensor (DCT concatenate in Figure~\ref{fig:preprocessing}). Lastly, every frequency channel is normalized by the mean and variance calculated from the training dataset.

The DCT reshape operation in Figure~\ref{fig:preprocessing} groups a two-dimensional DCT coefficients to a three-dimensional DCT cube. Since the JPEG compression standard uses $8\times8$ DCT transformation on the YCbCr color space, we group the components of the same frequency in all the  $8\times8$ blocks into one channel, maintaining their spatial relations at each frequency. Thus, each of the Y, Cb, and Cr components provides $8\times8=64$ channels, one for each frequency, with a total of $192$ channels in the frequency domain. Suppose the shape of the original RGB input image is $H\times W\times C$, where $C=3$ and the height and width of the image is denoted as $H$ and $W$, respectively. After converting to the frequency domain, the input feature shape becomes $H/8 \times W/8 \times 64C$, which maintains the same input data size. 

Since the input feature maps in the frequency domain are smaller in the $H$ and $W$ dimensions but larger in the $C$ dimension than the spatial-domain counterpart, we skip the input layer of a conventional CNN model, which is usually a stride-$2$ convolution. If a max-pooling operator immediately follows the input convolution (\textit{e.g.}, ResNet-50), we skip the max-pooling operator as well. Then we adjust the channel size of the next layer to match the number of channels in the frequency domain. It is illustrated in Figure~\ref{fig:connection}. This way, we  minimally adjust the existing CNN models to accept the frequency-domain features as input.

In the image classification task, the CNN models usually take input features of the shape $224 \times 224 \times 3$, which is usually downsampled from images with a much higher resolution. When the classification is performed in the frequency domain, larger images can be taken as input. Take ResNet-50 as an example, the input features in the frequency domain are connected to the first residue block with the number of channels adjusted to $192$, forming an input feature of the shape $56 \times 56 \times192$, as shown in Figure~\ref{fig:preprocessing}. That is DCT-transformed from input images of size $448\times 448\times 3$, which preserves four times more information than the $224 \times 224 \times 3$ counterpart in the spatial domain, at the cost of $4$ times the input feature size. Similarly, for the model MobileNetV2, the input feature shape is $112\times 112\times 192$, reshaped from images of size $896\times 896\times 3$. As discussed in Section~\ref{sec:static}, the majority of the frequency channels can be pruned without sacrificing accuracy. The frequency channel pruning operation is referred to as DCT channel select in Figure~\ref{fig:preprocessing}.


\begin{figure}[ht]
	\centering
	\includegraphics[width=0.46\textwidth]{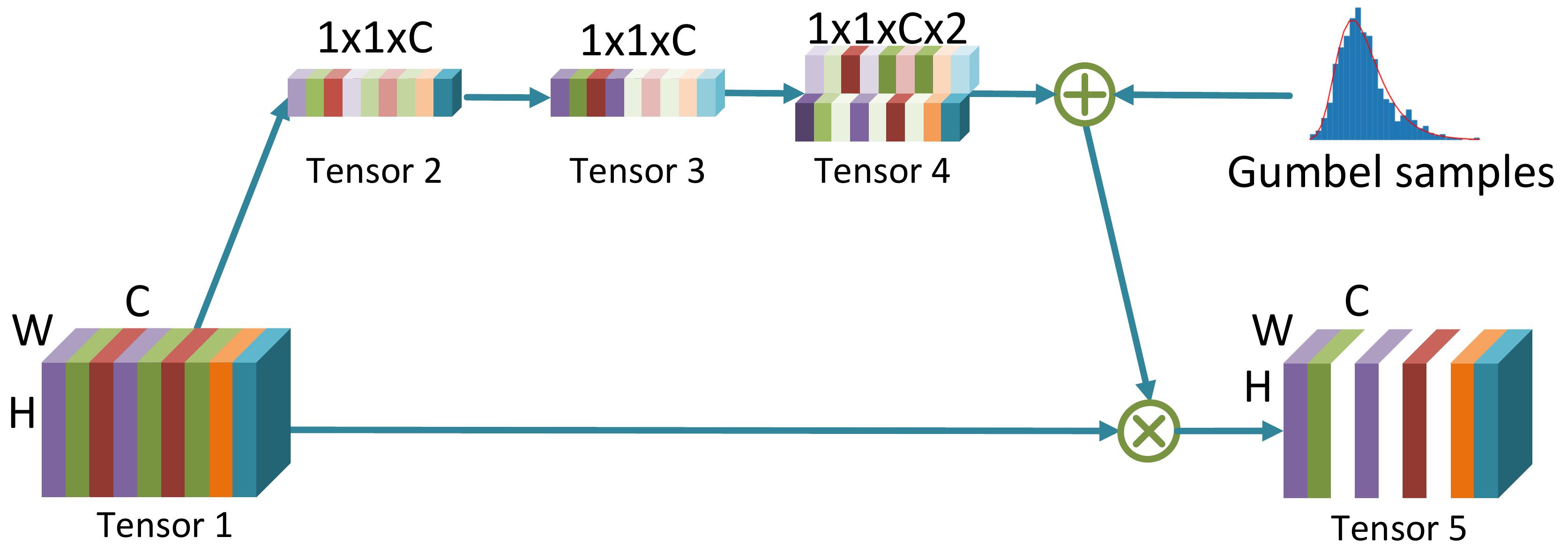}
	\caption{The gate module that generates the binary decisions based on the features extracted by the SE-Block. The white color channels of Tensor 5 indicate the unselected channels.}
	\label{fig:gate}
\end{figure}

\subsection{Learning-based Frequency Channel Selection} \label{sec:selection}
As different channels of the input feature are at different frequencies, 
we conjecture that some frequency channels are less informative to the subsequent tasks such as image classification, object detection, and instance segmentation, and removing the trivial frequency channels shall not result in performance degradation. Thus, we propose a learning-based channel selection mechanism to exploit the relative importance of each input frequency channel. 
We employ a dynamic gate module that assigns a binary score to each frequency channel. The salient channels are rated as one, the others as zero. The input frequency channels with zero scores are detached from the network. Thus, the input data size is reduced, leading to reduced computation complexity of domain transformation and communication bandwidth requirement. The proposed gate module is simple and can be part of the model to be applied in online inference.


Figure~\ref{fig:gate} describes our proposed gate module in detail. The input is of shape $W\times H\times C$ ($C=192$ in this paper), with $C$ frequency channels (Tensor 1 in Figure~\ref{fig:gate}). It is first converted to Tensor 2 in Figure~\ref{fig:gate} of shape $1\times1\times C$ by average pooling. Then it is converted to Tensor 3 in Figure~\ref{fig:gate} of shape $1\times1\times C$ by a $1\times 1$ convolutional layer. Conversion from Tensor 1 to Tensor 3 is exactly the same as a two-layer squeeze-and-excitation block (SE-Block) \cite{hu2018senet}, which utilizes the channel-wise information to emphasize the informative features and suppress the trivial ones. Then, Tensor 3 is converted to Tensor 4 in Figure~\ref{fig:gate} of the shape $1\times 1 \times C\times 2$ by multiplying every element in Tensor 3 with two trainable parameters. During inference, the two numbers for each of the $192$ channels in Tensor 4 are normalized and serve as the probability of being sampled as $0$ or $1$, and then, point-wise multiplied to the input frequency channels to obtain Tensor 5 in Figure~\ref{fig:gate}. As an example, if the two numbers in the $i$th channel in Tensor 4 are $7.5$ and $2.5$, there is a $75\%$ probability that the $i$th gate is turned off. In other words, the $i$th frequency channel in Tensor 5 becomes all zeros $75\%$ of the times, which effectively blocks this frequency channel from being used for inference.

Our gate module differs from the conventional SE-Block in two ways. First, the proposed gate module outputs a tensor of dimension $1\times 1\times C\times 2$, where the two numbers in the last dimension describe the probability of being on and off for each frequency channel, respectively. Thus we add another $1\times 1$ convolution layer for the conversion. Second, the number multiplied to each frequency channel is either $0$ or $1$, \textit{i.e.}, a binary decision of using the frequency or not. The decision is obtained by sampling a Bernoulli distribution $\textrm{Bern}(p)$, where $p$ is calculated by the $2$ numbers in the $1\times 1\times C\times 2$ tensor mentioned above.

One of the challenges in the proposed gate module is that the Bernoulli sampling process is not differentiable in case one needs to update the weights in the gate module. ~\cite{eric2017gumbel,tucker2017rebar, maddison2017concrete} propose a reparameterization method, called Gumbel Softmax trick, which allows the gradients to back propagate through a discrete sampling process (see Gumbel samples in Figure~\ref{fig:gate}).

Let $\boldsymbol{x}=(x_1,x_2,\ldots,x_C)$ be the input channels in the frequency domain ($C=192$) for a CNN model. 
Let  $\textbf{F}$ denote the proposed gate module such that $\textbf{F}(x_i)\in\{0,1\}$, for each frequency channel $x_i$. Then $x_i$ is selected if  
\begin{equation}
	\textbf{F}(x_i)\neq 0, \textrm{ i.e., } \textbf{F}(x_i) \odot x_i \neq \boldsymbol{0}, 
\end{equation}
where $\odot$ is the element-wise product. 


We add a regularization term to the loss function that balances the number of selected frequency channels, which is minimized together with the cross-entropy loss or other accuracy-related loss. Our loss function is thus as follows,
\begin{equation}
	\label{eq:block}
	\mathcal{L} = \mathcal{L}_{Acc} + \lambda \cdot \sum_{i=1}^C \textbf{F}(x_i), 
\end{equation}
where $\mathcal{L}_{Acc}$ is the loss that is related to accuracy. $\lambda$ is a hyperparameter indicating the relative weight of the regularization term. 


\begin{figure*}[ht]
	\centering
	\begin{subfigure}{0.63\linewidth}
		\centering
		\includegraphics[width=\textwidth]{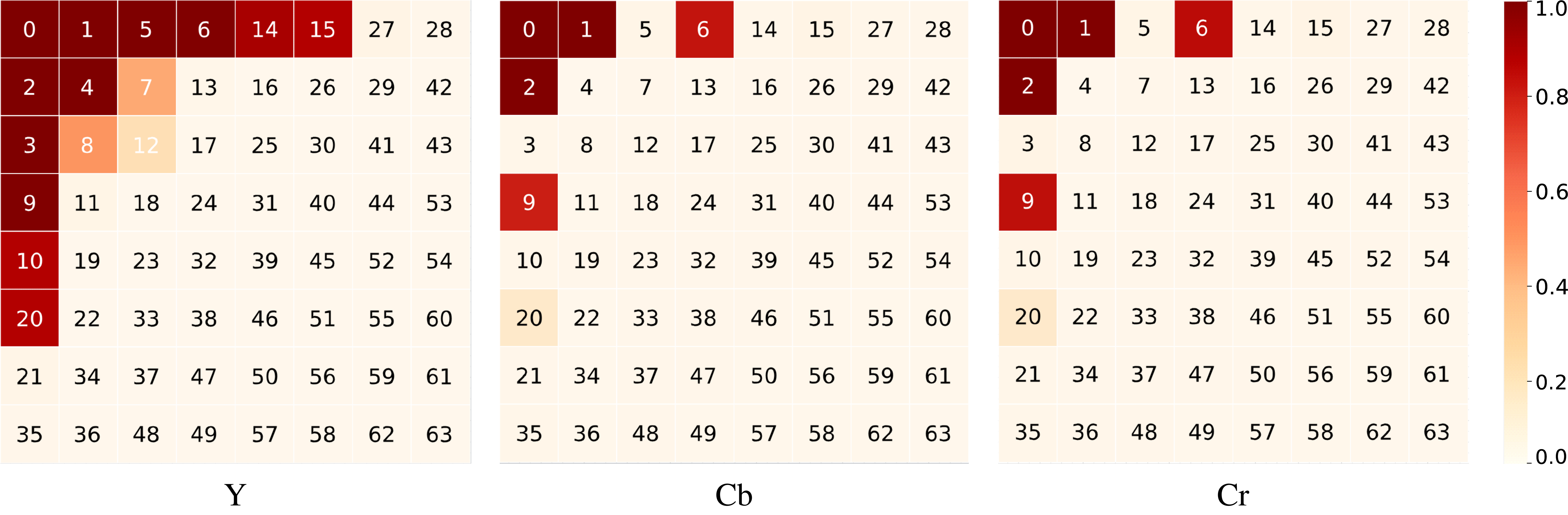}
		\caption{Heat maps of Y, Cb, and Cr components on the ImageNet validation dataset.}
		\label{fig:heatmap_cls}
	\end{subfigure}
	\\
	\begin{subfigure}{0.63\linewidth}
		\centering
		\includegraphics[width=\textwidth]{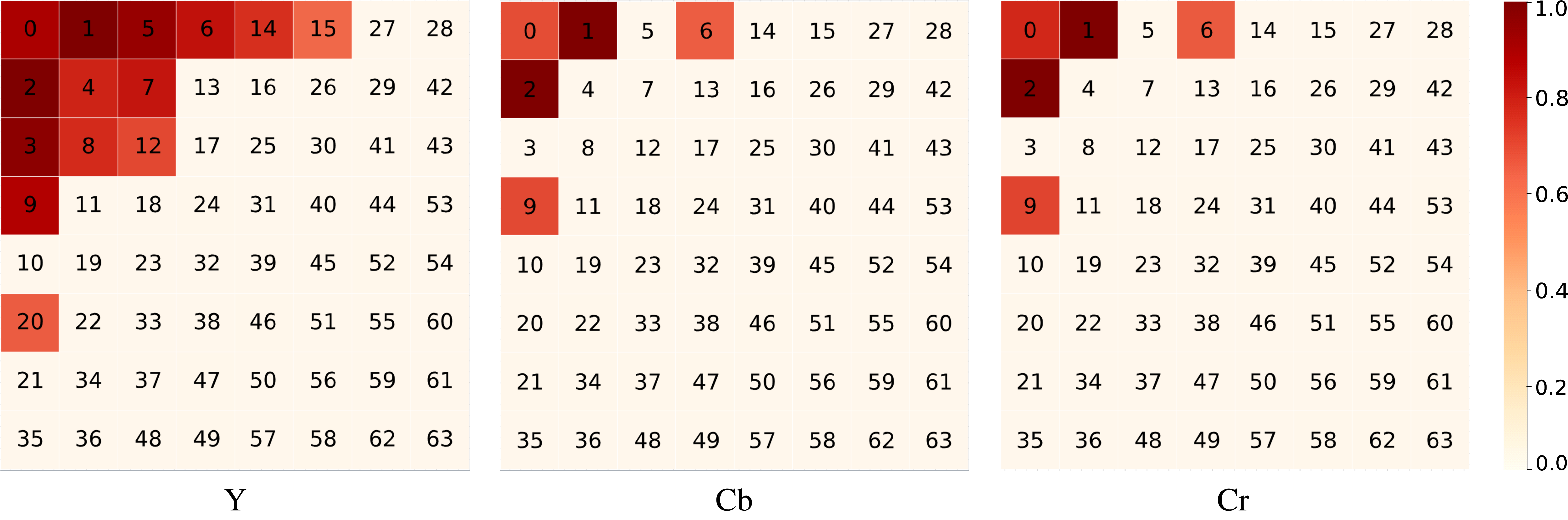}
		\caption{Heat maps of Y, Cb, and Cr components on the COCO validation dataset}
		\label{fig:heatmap_seg}
	\end{subfigure}
	\caption{A heat map  visualization of input frequency channels on the ImageNet validation dataset for image classification and  COCO validation dataset for instance segmentation. The numbers in each square represent the corresponding channel indices. The color from bright to dark indicates the possibility of a channel being selected from low to high. 
	}
	\label{fig:heatmap}
	\vspace{-2ex}
\end{figure*}

\subsection{Static Frequency Channel Selection} \label{sec:static}
The learning-based channel selection provides a dynamic estimation of the importance of each frequency channel, \textit{i.e.}, different input images may have different subsets of the frequency channels activated. 

To understand the pattern of frequency channel activation, we plot two heat maps, one on the classification task (Figure~\ref{fig:heatmap_cls}) and one on the segmentation task (Figure~\ref{fig:heatmap_seg}). The number in each box indicates the frequency index of the channel, with a lower and higher index indicating a lower and higher frequency, respectively. The heat map value indicates the likelihood a frequency channel being selected for inference across all the validation images.

Based on the patterns in the heat maps shown in Figure~\ref{fig:heatmap}, we make several observations:
\vspace{-1ex}
\begin{itemize}
	\itemsep0em 
	\item The low-frequency channels (boxes with small indices) are selected much more often than the high-frequency channels (boxes with with large indices). This demonstrates that low-frequency channels are more informative than high-frequency channels in general for vision inference tasks.
	\item
	The frequency channels in luma component Y are selected more often than the channels in chroma components Cb and Cr. This indicates that the luma component is more informative for vision inference tasks.
	\item
	The heat maps share a common pattern between the classification and segmentation tasks. This indicates that the above-mentioned two observations are not specific to one task and is very likely to be general to more high-level vision tasks.
	\item
	Interestingly, some lower frequency channels have lower probability of being selected than the slightly higher frequency channels. For example, in Cb and Cr components, both tasks favor Channel $6$ and $9$ over Channel $5$ and $3$.
\end{itemize}

Those observations imply that the CNN models may indeed exhibit similar characteristics to the HVS, and the image compression standards ({\it e.g.}, JPEG) targeting human eyes may be suitable for the CNN models as well.

The JPEG compression standard puts more bits to the low-frequency and the luma components. Following the same principle, we statically select the lower frequency channels, with more emphasis on the luma component than the chroma components. This ensures the frequency channels with higher activation probabilities are fed into the CNN models. The rest of the frequency channels can be pruned by either the image encoder or decoder to reduce the required data transmission bandwidth and input data size. 

\begin{table*}[ht]
	\centering
	\caption{ResNet-50 classification results on ImageNet (validation). The input size of each method is normalized over the baseline ResNet-50. The input frequency channels are selected with the square and triangle channel selection pattern if the postfix S and T is specified, respectively.}
	\scalebox{0.75}{		
		\begin{tabular}{c c c | c c c}
			\toprule [0.2em]
			ResNet-50 & \#Channels & Size Per Channel & Top-1  & Top-5 & Normalized Input Size  \\ \hline
			RGB & 3 & 224$\times$224 & 75.780 & 92.650 & 1.0 \\
			YCbCr & 3 & 224$\times$224 & 75.234 & 92.544 & 1.0 \\ \hline
			DCT-192~\cite{gueguen2018jpeg} & 192 & 28$\times$28 & 76.060 & 93.020 & 1.0 \\ 
			\textbf{DCT-192 (ours)} & 192 & 56$\times$56 & 77.194 & 93.454 & 4.0       \\ \hline
			\textbf{DCT-24D (ours)} & 24 & 56$\times$56 & 77.166 & 93.560 & 0.5  	\\ 	
			\textbf{DCT-24S (ours)} & 24 & 56$\times$56 & 77.196 & 93.504 & 0.5  	\\
			\textbf{DCT-24T (ours)} & 24 & 56$\times$56 & 77.148 & 93.326 & 0.5  	\\ \hline
			
			\textbf{DCT-48S (ours)} & 48 & 56$\times$56 & 77.384 & 93.554 & 1.0       \\
			\textbf{DCT-48T (ours)} & 48 & 56$\times$56 & 77.338 & 93.614 & 1.0       \\ \hline
			\textbf{DCT-64S (ours)} & 64 & 56$\times$56 & 77.232 & 93.624 & 1.3       \\
			\textbf{DCT-64T (ours)} & 64 & 56$\times$56 & 77.280 & 93.456 & 1.3       \\ 
			\bottomrule[0.2em]
		\end{tabular}
	}
	\label{tab:resnet_classification}
\end{table*}

\begin{table*}[h]
	\centering
	\caption{MobileNetV2 classification results on ImageNet (validation).}
	\label{tab:mobilenet_classification}
	\scalebox{0.75}{
		\begin{tabular}{c c c | c c c}
			\toprule [0.2em]
			MobileNetV2& \#Channels & Size Per Channel  & Top-1  & Top-5 & Normalized Input Size \\ \hline
			RGB        & 3        & 224$\times$224  & 71.702 & 90.415 & 1.0 \\ \hline
			\textbf{DCT-6S (ours)}  & 6  & 112$\times$112 & 71.776 & 90.258 & 0.5 \\
			\textbf{DCT-12S (ours)} & 12 & 112$\times$112 & 72.156 & 90.634 & 1.0 \\
			\textbf{DCT-24S (ours)} & 24 & 112$\times$112 & 72.364 & 90.606 & 2.0 \\			
			\textbf{DCT-32S (ours)} & 32 & 112$\times$112 & 72.282 & 90.592 & 2.7 \\
			\bottomrule[0.2em]
		\end{tabular}
	}
	\vspace{-2ex}
\end{table*}

\section{Experiment Results}
We benchmark our proposed methodology on three different high-level vision tasks: image classification, detection, and segmentation.

\subsection{Experiment Settings on Image Classification} \label{sec:cls}
We benchmark our method on image classification using the ImageNet 2012 Large-Scale Visual Recognition Challenge dataset (ILSVRC-2012)~\cite{deng2009imagenet}. We use the stochastic gradient descent (SGD) optimizer. SGD is applied with an initial learning rate of $0.1$, a momentum of $0.9$, and a weight decay of 4e-5. We choose ResNet-50~\cite{he2017deep} and MobileNetV2~\cite{sandler2018mobilenetv2} as the CNN models because they contain important building blocks (\textit{e.g.}, residue blocks and depthwise separable convolutions) widely used in modern CNN models. Note that our method can be generally applied to any CNN model. We train $210$ and $150$ epochs and decay the learning rate by $0.1$ every $50$ epochs for ResNet-50 and MobileNetV2, respectively.

To normalize the input channels, we compute the mean and variance of the DCT coefficients for each of the 192 frequency channels separately on all the training images. 

As described in Section~\ref{sec:preprocess}, the input features in the frequency domain are generated from images with a much higher resolution than the spatial-domain counterpart. However, some of the images in the ImageNet dataset have lower resolutions. We perform similar pre-processing steps as in the spatial domain, including resizing and cropping to a larger image size, performing upsampling when needed. 

\subsection{Experiment Results on Image Classification}
We train the ResNet-50 model with $192$ frequency channel inputs on the image classification task using the approach described in Section~\ref{sec:selection}. The gate module for channel selection is trained together with the ResNet-50 model. Figure~\ref{fig:heatmap_cls} shows a heat map of the selection results over the validation set with $\lambda=0.1$. Note that different regularization parameters $\lambda$ generate different number of activated frequency channels in heat maps. A typical example is shown in Figure~\ref{fig:heatmap_cls}, that most channels ($\geq 80\%$) have very low possibility ($\leq 3\%$) of being selected. 

Observing that low frequency channels are more important in the heat maps, we explore the sensitivity of the precise shapes of selected channels. In Table~\ref{tab:resnet_classification}, DCT-24D shows the accuracy when 24 (14+5+5) channels are precisely selected based on the result of the dynamic selection in Figure~\ref{fig:heatmap_cls}. In comparison, DCT-24T and DCT-24S show the accuracy when a total of 24 channels for Y, Cb, Cr components are close to upper-left triangles and squares, respectively. 
The variation of the top-1 accuracy is almost negligible and all of them outperform a baseline ResNet-50 by roughly $1.4\%$. 
This demonstrates that the benefit of the proposed frequency-domain learning can be applied to many tasks as long as a majority of low-frequency channels are selected.
Note the input data size is only a half of the baseline ResNet-50. Since DCT-24S provides a slightly better result, the remaining static selection are based on patterns that are close to upper-left squares (some lower right channels may be missing).

Similarly, we choose the top $(32,8,8)$ channels for DCT-48S/T and top $(44,10,10)$ channels for DCT-64S/T. The results on the ImageNet dataset are shown in Table~\ref{tab:resnet_classification} along with selecting all $192$ frequency channels.
In particular, compared with the baseline ResNet-50, the top-1 accuracy is improved by $1.4\%$ using all frequency channels. 
One should also note that the accuracy is dropped when the inputs are transformed from the RGB to the YCbCr color space (both in the spatial domain) by roughly $0.5\%$, and the improvement of our method (in the frequency domain) over the YCbCr case is even larger.


Another interesting observation is that the model trained with a subset of channels may perform better than the model trained with all the 192 channels. Such a counter-intuitive observation implies that a small number (e.g., 24) of low-frequency channels are sufficient to capture useful features and additional frequency components may introduce noise. 


Similar experiments are performed using the MobileNetV2 as the baseline CNN model and the results are shown in Table~\ref{tab:mobilenet_classification}.
Note that DCT-12S and DCT-6S select $12$ and $6$ frequency channels, and the input data size is the same and a half of the baseline MobileNetV2, respectively. The top-1 accuracy of DCT-12S and DCT-6S is improved by $0.454\%$ and $0.074\%$, respectively. The top-1 accuracy is improved by $0.662\%$ and $0.580\%$ by selecting $32$ and $24$ frequency channels, respectively. 


\begin{table*}[htbp]
	\centering
	\caption{Bbox AP results of Mask R-CNN using different backbones on COCO 2017 validation set. The baseline Mask R-CNN uses a ResNet-50-FPN as the backbone. The DCT method uses the frequency-domain ResNet-50-FPN as the backbone.}
	\scalebox{0.75}{
		\begin{tabular}{ccc|cccccc}
			\toprule [0.2em]
			\multirow{2}{*}{Backbone} & \multicolumn{1}{l}{\multirow{2}{*}{\#Channels}} & \multirow{2}{*}{Size Per Channel} & \multicolumn{6}{c}{bbox}                    \\ \cline{4-9} 
			& \multicolumn{1}{l}{}                          &                           & AP   & AP@0.5 & AP@0.75 & AP$_\text{S}$  & AP$_\text{M}$  & AP$_\text{L}$  \\ \hline
			ResNet-50-FPN (RGB)                & 3         & 800$\times$1333                & 37.3 & 59.0   & 40.2    & 21.9 & 40.9 & 48.1 \\ \hline
			\textbf{DCT-24S (ours)}             & 24       & 200$\times$334               & 37.7 & 59.2   & 40.9    & 21.7 & 41.4 & 49.1 \\
			\textbf{DCT-48S (ours)}             & 48       & 200$\times$334  & 38.1 & 59.5 & 41.2 & 22.0 & 41.3 & 49.8 \\
			\textbf{DCT-64S (ours)}             & 64       & 200$\times$334                & 38.1 & 59.6   & 41.1    & 22.5 & 41.6 & 49.7 \\

			\bottomrule[0.2em]
		\end{tabular}
	}
	\label{tab:detection}
	\vspace{-1ex}
\end{table*}

\begin{table*}[htbp]
	\centering
	\caption{Mask AP results of Mask R-CNN using different backbones on COCO 2017 validation set.}
	\scalebox{0.75}{
		\begin{tabular}{ccc|cccccc}
			\toprule [0.2em]
			\multirow{2}{*}{Backbone} & \multicolumn{1}{l}{\multirow{2}{*}{\#Channels}} & \multirow{2}{*}{Size Per Channel} & \multicolumn{6}{c}{mask}                    \\ \cline{4-9} 
			& \multicolumn{1}{l}{}                          &                           & AP   & AP@0.5 & AP@0.75 & AP$_\text{S}$  & AP$_\text{M}$  & AP$_\text{L}$  \\ \hline
			ResNet-50-FPN (RGB) & 3 & 800$\times$1333 & 34.2 & 55.9 & 36.2 & 15.8 & 36.9 & 50.1 \\ \hline
			\textbf{DCT-24S (ours)} & 24 & 200$\times$334 & 34.6 & 56.1 & 36.9 & 16.1 & 37.4 & 50.7 \\
			\textbf{DCT-48S (ours)} & 48 & 200$\times$334 & 35.0 & 56.6 & 37.2 & 16.3 & 37.5 & 52.3  \\
			\textbf{DCT-64S (ours)} & 64 & 200$\times$334 & 35.0 & 56.5 & 37.4 & 16.9 & 37.6 & 51.6 \\
			\bottomrule[0.2em]
		\end{tabular}
	}
	\label{tab:segmentation}
	\vspace{-1ex}
\end{table*}

\subsection{Experiment Settings on Instance Segmentation}
We train our model on the COCO train2017 split containing about 118k images and evaluate on the val2017 split containing 5k images. We evaluate the bounding box (bbox) average precision (AP) for the object detection task and the mask AP for the instance segmentation task. Based on the Mask R-CNN~\cite{he2017mask}, our model consists of a frequency-domain ResNet-50 model as introduced in Section~\ref{sec:cls} and a feature pyramid network \cite{lin2017fpn} as the backbone. The frequency-domain ResNet-50 model is fine-tuned with the bounding-box recognition head and the mask prediction head. Input images are resized to a maximum scale of $1600\times2666$ without changing the aspect ratio. The corresponding DCT coefficients have a maximum size of $200\times334$, which are fed into the ResNet-50-FPN \cite{lin2017fpn} for feature extraction.

We train our networks for $20$ epochs with an initial learning rate of $0.0025$, which is decreased by $10\times$ after $16$ and $19$ epochs. The rest of the configurations follow those of MMDetection~\cite{chen2019mmdetection}.

In Table~\ref{tab:detection} and Table~\ref{tab:segmentation}, we report the AP metric that averages APs across IoU thresholds from $0.5$ to $0.95$ with an interval of $0.05$. Both the bbox AP and the mask AP are evaluated. For the mask AP, we also report AP@0.5 and AP@0.75 at the IoU threshold of $0.5$ and $0.75$ respectively, as well as AP$_S$, AP$_M$, and AP$_L$ at different scales. 

\subsection{Experiment Results on Instance Segmentation}
We train our Mask R-CNN model using the 192-channel inputs in the frequency domain for instance segmentation. The gate module for dynamic channel selection is trained together with the entire Mask R-CNN. Figure~\ref{fig:heatmap_seg} shows the heat maps for the dynamic selection. 

We further train our models using only the top $24$, $48$, and $64$ high-probability frequency channels. The bbox and mask AP of our method in different cases is reported in Table~\ref{tab:detection} and Table~\ref{tab:segmentation}, respectively. The experiment results show that our method outperforms the RGB-based Mask R-CNN baseline with both an equal (DCT-48S) or smaller (DCT-24S) input data size. Specifically, the 24-channel model (DCT-24S) achieves an improvement of $0.4$ in both bbox AP and mask AP with a half of the input data size compared to the RGB-based Mask R-CNN baseline.

\begin{figure}[ht]
	\centering
	\includegraphics[width=0.48\textwidth]{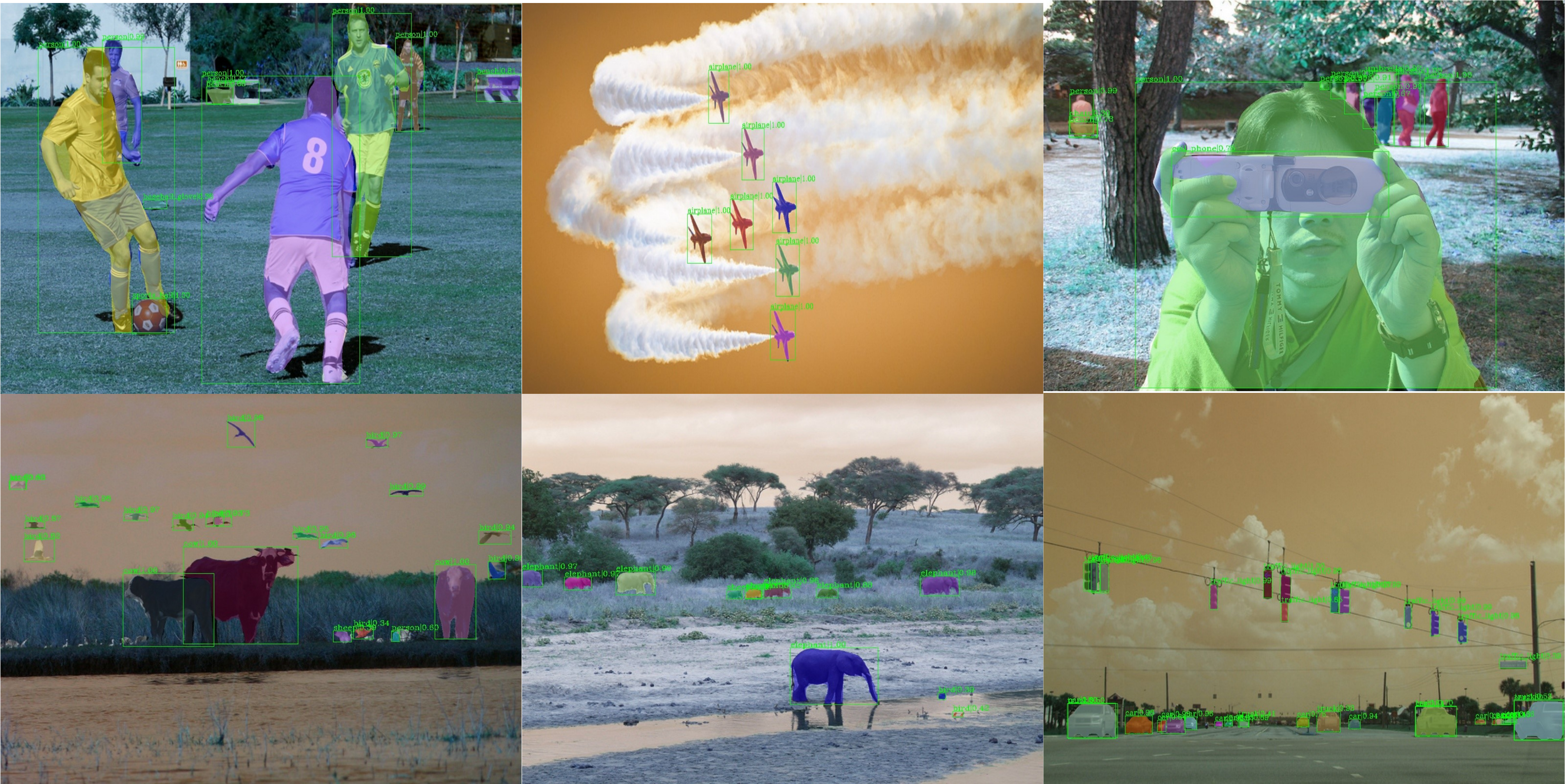}
	\caption{Examples of instance segmentation results on the COCO dataset.}
	\label{fig:seg_visual}
	\vspace{-3ex}
\end{figure}

Figure~\ref{fig:seg_visual} visually illustrates the segmentation results of the Mask R-CNN model trained and performing inference in the frequency domain.

\section{Conclusion}
In this paper, we propose a method of learning in the frequency domain and demonstrate its generality and superiority for a variety of tasks, including classification, detection, and segmentation. Our method requires little modification to the existing CNN models that take RGB input thus can be generally applied to existing network training and inference methods.
We show that learning in the frequency domain better preserves image information in the pre-processing stage than the conventional spatial downsampling approach and consequently achieves improved accuracy. We propose a learning-based dynamic channel selection method and empirically show that the CNN models are more sensitive to low-frequency channels than high-frequency channels. Experiment results show that one can prune up to $87.5\%$ of the frequency channels using the proposed channel selection method with no or little accuracy degradation in the classification, object detection, and instance segmentation tasks.

\textbf{Acknowledgement.} The work by Arizona State University is supported by an NSF grant (IIS/CPS-1652038). 


\clearpage


\pretitle{\vspace{10ex}}
\posttitle{\vspace{10ex}}
\title{\textit{Supplementary Material for} \\ Learning in the Frequency Domain}
\preauthor{}
\postauthor{}
\author{}

\maketitle

\renewcommand{\thesection}{\Alph{section}}
\setcounter{section}{0}


This document supplements our paper entitled Learning in the Frequency Domain by providing further quantitative and qualitative insights of the results. 

\section{Instructions to Reproduce the Experiments}
We have provided the source code to reproduce the experiments in the paper. The code is based on PyTorch and is available at \url{https://github.com/calmevtime/DCTNet}. There are two folders in the repo named ``classification" \footnote{\url{https://github.com/calmevtime1990/supp/tree/master/classification}} and ``segmentation" \footnote{\url{https://github.com/calmevtime1990/supp/tree/master/segmentation}}. The classification folder contains all the necessary code and instructions to reproduce our work using the pretrained models on the image classification task. The segmentation folder contains all the necessary code and instructions on the object detection and instance segmentation task.

\section{Additional Instance Segmentation Results}
\begin{figure*} [t]
	\centering
	\includegraphics[width=\textwidth]{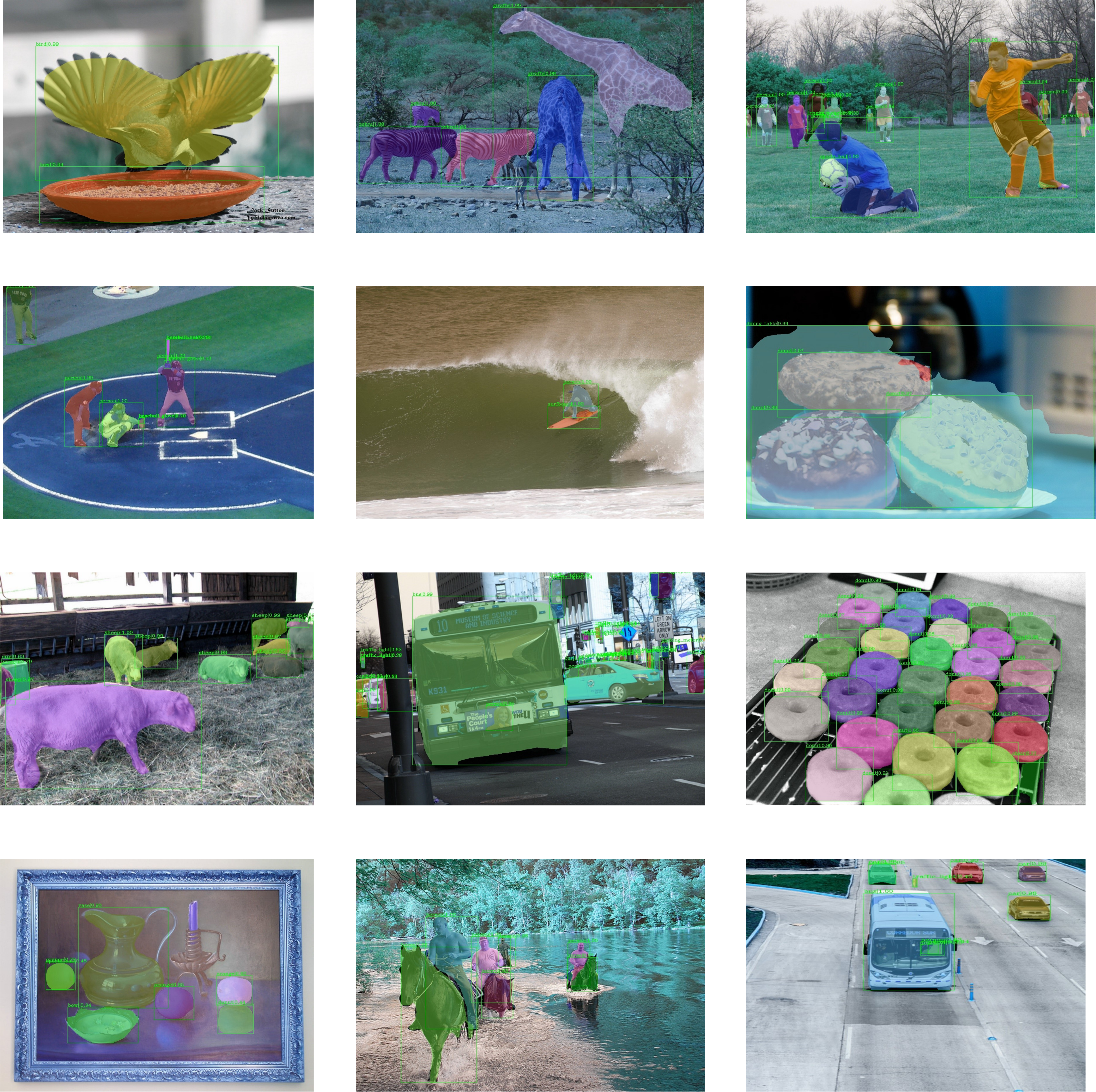}
	\caption{Examples of instance segmentation results on the COCO dataset.}
	\label{fig:visual_examples}
\end{figure*}

More instance segmentation examples are shown in Figure~\ref{fig:visual_examples}.

\section{Object Detection Results on Faster R-CNN}
In addition to the Mask R-CNN model provided in the paper, we train our model for object detection on the COCO train2017 split and evaluate on the val2017 split using  the  Faster  R-CNN \cite{ren2017faster} model. Our  model  consists of a frequency-domain ResNet-50 model (introduced in Section 4.1 in the main paper) and a feature pyramid network \cite{lin2017fpn} as the backbone. The frequency-domain ResNet-50 model is fine-tuned with the classification head and bounding box regression head. Input images are resized to a maximum scale of1600$\times$2666 without changing the aspect ratio. The corresponding DCT coefficients have a maximum size of 200$\times$334, which are fed into the ResNet-50-FPN for feature extraction. The rest of the configurations follow those of MMDetection \cite{chen2019mmdetection}.

In Table~\ref{tab:detection}, we report the results on the object detection task using the frequency domain Faster R-CNN. The proposed method achieves a 0.8$\%$ AP improvement compared to the baseline Faster R-CNN on the COCO dataset.

\begin{table*}[]
	\centering
	\caption{Bbox AP results of Faster R-CNN using different backbones on COCO 2017 validation set. The baseline Mask R-CNN use a ResNet-50-FPN as the backbone.  The DCT method uses the frequency-domain ResNet-50-FPN as the backbone.}
	\label{tab:detection}
	\scalebox{1.0}{
		\begin{tabular}{ccc|cccccc}
			\toprule [0.2em]
			\multirow{2}{*}{Backbone} & \multicolumn{1}{l}{\multirow{2}{*}{\#Channels}} & \multirow{2}{*}{Size Per Channel} & \multicolumn{6}{c}{bbox}                    \\ \cline{4-9} 
			& \multicolumn{1}{l}{}                          &                           & AP   & AP@0.5 & AP@0.75 & AP$_\text{S}$  & AP$_\text{M}$  & AP$_\text{L}$  \\ \hline
			ResNet-50-FPN (RGB)                & 3         & 800$\times$1333                & 36.4 & 58.4   & 39.1    & 21.5 & 40.0 & 46.6 \\ \hline
			\textbf{DCT-24 (ours)}             & 24       & 200$\times$334               & 37.2 & 58.8   & 39.9    & 21.9 & 40.7 & 48.9 \\
			\textbf{DCT-48 (ours)}             & 48       & 200$\times$334  & 37.1 & 58.6  & 40.2 & 21.7 & 40.9 & 48.8 \\
			\textbf{DCT-64 (ours)}             & 64       & 200$\times$334                & 37.2 & 58.5   & 40.6    & 21.9 & 40.9 & 48.3 \\
			\bottomrule[0.2em]
		\end{tabular}
	}
\end{table*}

\end{document}